%% file: main.tex
\documentclass[sigconf,nonacm]{acmart}
\input{glyphtounicode}
\usepackage{tikz}
\usetikzlibrary{positioning,arrows.meta}
\usepackage{float}
\usepackage{multirow}
\usepackage{subcaption}
\usepackage{fancyvrb}

\newcommand{\ms}[2]{$#1_{\,\pm#2}$}
\newcommand{\msb}[2]{$\mathbf{#1}_{\,\pm#2}$}

\begin{document}

\title{Q\&A on Any Spreadsheet Requires Interpreting Its Grid Structure}

%% acmart on TeX Live 2025 (arXiv default) REQUIRES \country inside \affiliation.
\author{Zofia Smole\'{n}}
\email{zofsmolen@gmail.com}
\orcid{0009-0001-4722-8809}
\affiliation{%
  \institution{Systems Research Institute, Polish Academy of Sciences}
  \city{Warsaw}
  \country{Poland}
}

\begin{abstract}
Semantic cell annotation improves chunking interpretability for spreadsheets in LLM-driven RAG systems, aiding answer generation through enriched context rather than improved retrieval accuracy. We propose a novel framework of splitting any spreadsheet into interpretable chunks using cell role annotation. Our framework beats the state of the art, yet it faces a hard ceiling. Spreadsheets are fundamentally two-dimensional unstructured data with continuous relationships and infinite potential cell roles. Because classification models are restricted to finite, pre-defined classes, they cannot perfectly capture this structural nuance---even with human-level annotation. We show that addressing the spreadsheet-to-LLM bottleneck requires moving beyond discrete cell classification. Instead, the field must develop dimensionality-reduction techniques to directly flatten 2D unstructured spreadsheets into 1D unstructured text. Text chunks would be easier for downstream RAG to interpret and generate from.
\end{abstract}


\begin{teaserfigure}
  \centering
  \includegraphics[width=\textwidth]{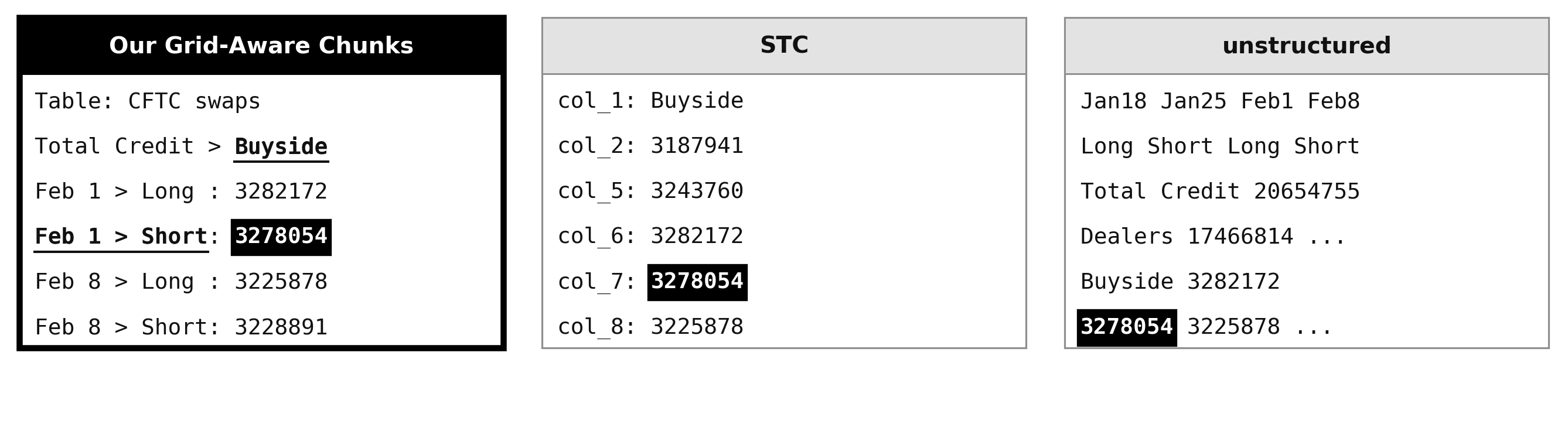}
  \caption{The same answer cell (highlighted) as it reaches the LLM in chunks built by three methods, on a real nested-header sheet from our corpus (a CFTC swaps report). Unstructured emits bare rows with the headers detached; STC maps values to generic first-row column labels that name nothing. Our grid-aware Row chunk spells out the full context---sheet title and the complete header path (Total Credit\,$\rightarrow$\,Buyside, Feb\,1\,$\rightarrow$\,Short)---so the generator can interpret the number without guessing.}
  \label{fig:teaser}
\end{teaserfigure}

\maketitle

\section{Introduction}

\begin{figure}[t]
  \centering
  \includegraphics[width=0.8\linewidth]{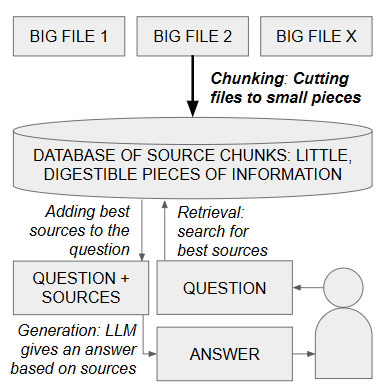}
  \caption{Retrieval-Augmented Generation. This paper is about the first step---chunking---for spreadsheets: how the sheet is cut determines what context each retrieved value carries into generation.}
  \label{fig:rag}
\end{figure}

One of the most common applications of language models is answering questions. Very often the question asked cannot be answered just based on data the model was trained on---for example when the information requested isn't public (like undisclosed financial data or company secrets). In such situations the user can provide context to the language model by putting it in the same prompt as the question. If the source with the information requested is small enough, it can simply be stuffed into the prompt as a whole. However, given limited context windows of language models, putting a 5-year history of customer claims against Walmart in the prompt would be difficult if not impossible. A standard workaround is Retrieval-Augmented Generation (RAG): first dividing the context into small pieces (chunking), then searching for the pieces most relevant to the query (retrieval), and at the end generating the answer based on the question asked and the chunks retrieved (Figure~\ref{fig:rag}).

In this paper we focus on creating those chunks of knowledge, called chunking. It is comparatively easy to chunk text or other sequential data: it is usually enough to split on patterns such as a period, a comma, a formatting type (headline, paragraph), or even a fixed number of characters. However, chunking unstructured data organised in some way other than sequentially is still a research gap. In particular, very little has been done to efficiently chunk real-life Excel spreadsheets. Tables are often associated with structured data sources, but in this paper we look broadly at spreadsheets: real-life sheets (often created in MS Excel or Google Sheets) can contain tables in any format and quantity on a single tab. Extracting data from \emph{any} spreadsheet---not a specific spreadsheet created in a known format---requires upfront interpretation of its structure. We propose a novel framework where we first semantically annotate cell roles and then use these roles to extract meaningful, grid-aware chunks.

Frameworks currently most used in industry mostly just cut XML into pieces. This strips extracted values of their structural context: column, row, and table headers among others. Chunks become blurbs of text and values that happen to be visually close to each other, and such chunks are usually useless. The difficulty is confirmed in practice: across 59 active practitioner discussions on machine-learning forums (r/Rag, r/LangChain) we scraped, the most-cited failure mode was loss of context during chunking (59\% of threads), followed by complex or nested headers (37\%) and messy, sparse layouts (29\%). Practitioners report a fragmented toolchain---49\% iterate rows with Pandas, 32\% use Text-to-SQL agents (which translate a question into a database query), 25\% resort to vision LLMs to find relationships in the data, 22\% use Docling~\cite{docling2024} (an open-source document-parsing toolkit whose Excel backend detects the table regions on a sheet), 17\% use Unstructured.io,\footnote{\url{https://unstructured.io}} and 12\% parse HTML with BeautifulSoup.\footnote{\url{https://www.crummy.com/software/BeautifulSoup/}} None of these reliably preserves the header$\rightarrow$value hierarchy that carries most of the context.

Research on efficiently extracting spreadsheet data for RAG is limited, but two works stand out: SpreadsheetLLM with its Chain of Spreadsheet~\cite{dong2024spreadsheetllm}, and STC~\cite{stc2026}. SpreadsheetLLM starts from a simple problem: spreadsheets are too big for an AI to read in one go. So it shrinks them---about 25 times smaller. It does this by keeping only the rows and columns where something changes (like where a table or its headers begin), by writing each repeated value once instead of fifty times, and by replacing long runs of numbers with a short note like ``these cells are all whole numbers''. What is left is a small map of the sheet: you can see where everything is, but the actual numbers have been thrown out.

A map with no numbers cannot answer a question, and that is why the authors provide a retrieval method called Chain of Spreadsheet. An LLM reads the map and the question, and instead of answering, it points: ``the answer lives around B2:D40''. That small piece is then cut out of the original sheet---with the real numbers still in it---and given to the LLM again, and now it answers. The weak point is that this assumes you already know which file to look in. When there are hundreds of sheets, something has to find the right one first---and all there is to search through are those shrunken maps, which are hard to match against a question because the numbers are gone.

STC then proved that guiding chunk extraction with cell roles can improve the quality of answers generated with RAG, though it does not explain why. It slices a spreadsheet into retrieval chunks with a fixed rule: it treats the first populated row as the header, emits one key--value block per data row, and merges those blocks up to a token budget. It assumes every sheet has the same shape---header on top, data below---and stamping that template onto a sheet of that shape yields context-rich chunks. Unfortunately, not every table looks like that; but such an approach, if generalized, could be more exact than using an LLM-generated map.

We build on the idea of using cell roles for chunk extraction, but we do not assume a fixed spreadsheet structure. Such generalization requires \emph{learning} cell roles and the relationships between them---a cell annotation task with a long history. The earliest attempts relied on hand-written rules: if a cell is bold, sits in the first row, or is followed by a column of numbers, call it a header---that kind of logic, encoded by hand (e.g., the DeExcelerator pipeline~\cite{eberius2013deexcelerator}). Rules like these work on tidy sheets and break on everything else, much the same weakness that STC's fixed template runs into today. The next wave let the machine learn the rules instead: Fang et al.~\cite{fang2012header} trained a feature-based classifier to detect and classify table headers in document tables; Chen and Cafarella~\cite{chen2013automatic} taught a statistical model to label whole rows of web-published spreadsheets as titles, headers or data; Koci and colleagues~\cite{koci2016layout} pushed the granularity down to the single cell, feeding a classifier dozens of hand-picked clues---the cell's formatting, its content type, where it sits in the grid. Their DECO corpus~\cite{koci2019deco} of real, manually annotated spreadsheets gave the field shared training data, though with only a handful of coarse roles along the lines of data, header, derived value and note. The third wave dropped the hand-picked clues too and let neural networks read the grid directly: learned cell embeddings, multi-task networks that extract header structure straight from the sheet, TabularNet~\cite{du2021tabularnet} combining a recurrent network with a graph network over neighbouring cells, and large-scale pretraining on millions of tables (TUTA~\cite{wang2021tuta}). The trend line is clear---from rules, to learned decisions over designed features, to learned features over the raw grid, with each step handling messier spreadsheets than the last.

What all of these share, however, is that the cell label is the finish line: models are scored on classification accuracy and the story ends there. For chunking we need two things more. First, finer labels---it is not enough to know that a cell is ``a header''; we need to know how deep it sits in the header hierarchy, because a level-two header describes a different slice of the data than a level-one header, and a chunk must carry the right one. Second, we need the labels to prove their worth downstream: the test of a cell role is not whether it matches an annotation, but whether the chunk built from it lets a model answer a question. Those two requirements---depth-aware roles, judged by the answers they enable---are where our work picks up.

To learn cell roles we train six neural networks from two different families. Three of them look at each cell on its own---its text and its formatting---and guess the role. The other three also learn how cells are connected, treating the sheet as a graph where neighbouring cells share information. The two families read a spreadsheet in very different ways, and that is on purpose: if both make RAG better, then the credit goes to the idea itself, not to one lucky network. That is what happens. On our benchmark of 480 questions, chunks built from learned roles beat every baseline we test: the best model scores 3.88 out of 5 on human-rated answer quality; the strongest competitor gets 3.43.

We can also say why it works. Surprisingly, good roles do not help the search step much---a chunk matches a question through its words, whether the roles are right or not. The roles pay off after retrieval, when the model writes the answer. A model that sees ``51.5'' next to the words that describe it can use the number. A model that sees a bare ``51.5'' cannot. In short: structure detection does not make the right chunk easier to find, it makes the found chunk easier to understand.

Finally, we check how far this approach can go. We rebuild the chunks using roles annotated by a human---perfect structure understanding---and the score rises only a little, to 4.01. So the approach has a ceiling, and even perfect roles sit far from a perfect score. Some spreadsheets are simply hard. And on the simplest ones, our machinery adds nothing: a plain first-row-header rule does just as well. The real bottleneck is no longer recognising the structure. It is deciding how to cut the sheet into chunks---because no single cutting rule works for every sheet. That is why we end by suggesting a less structured way of building chunks: one that adapts to each sheet, possibly even turning records into plain sentences, instead of forcing one shape on everything.

\section{Methodology}

Our methodology proceeds in three steps. First, we train six cell-role annotation models from two families---node classifiers and graph learners---on human-labelled spreadsheets. Second, we test whether the learned roles help downstream Q\&A. We build chunks from each model's predictions and compare them, on retrieval (recall@1/@5) and on a 1--5 answer rating from a human judge, against five baselines: the two most popular Python ingestion paths (BeautifulSoup and Unstructured) and three academic approaches (STC, STC with Docling's table split, and SpreadsheetLLM's SheetCompressor with Chain-of-Spreadsheet adapted to multiple sheets). We also compare against chunks built from human-annotated gold roles, which give the ceiling of the approach. Our models beat all baselines, but even the gold-role ceiling stays well short of a 5/5 rating. Third, we ask why the approach wins and where its limit lies. Regressing both metrics on role-model F1 across sixteen deployed checkpoints shows that role quality drives generation, not retrieval. An ablation that removes one role at a time from the gold chunks shows the value is carried by the header roles. And a breakdown by sheet structure shows the gains concentrate on complicated layouts---nested headers, cross-tabs, multiple tables---while on simple flat tables, where a first-row header already names every column, all methods converge to a tie.

\subsection{Dataset}

For training, we used 505 spreadsheet tabs from the Sheetpedia corpus~\cite{sheetpedia2024}, each with cell roles annotated by two human labelers (with peer review) directly in native Excel. Together these sheets contain about 1.0M annotated cells, roughly 550K of them non-empty. Of the 505 tabs, 496 pass minimal-size filtering and form the training and evaluation pool for the node classifiers; the graph learners use the 419-sheet subset with multi-table structure. Since the two families are evaluated on slightly different pools, their macro-F1 scores are not perfectly cross-comparable.

For RAG evaluation, we used 480 questions targeting 80 held-out sheets, embedded in a corpus with 302 additional distractor sheets (382 in total), so retrieval has to find the right sheet among many plausible ones. The 80 answer sheets also carry full human role annotations, which serve as a ceiling estimate: from these gold roles we can build perfect chunks and measure the best a role-based method could do if it recognized every cell role correctly. Every answer generated from chunks---ours and every baseline's---was rated 1--5 by a human judge.

Each answer sheet was also tagged with one or more structural categories, which we use to break results down by layout difficulty. A \emph{simple flat table} has one header row at the top and plain value rows below---the layout most tools assume. \emph{Nested headers} means the headers form a hierarchy (e.g., Year split into Q1--Q4), so a value only makes sense with its full header path. A \emph{matrix / cross-tab} indexes each value by a row header and a column header at once (e.g., regions down the side, years across the top). A \emph{multi-table} sheet holds several separate tables, so their boundaries must be found before anything else can be read. The categories can overlap. Of the 480 questions, 81 target nested headers, 200 matrix/cross-tabs, 86 multi-table sheets, and 99 simple flat tables.

Annotation uses 13 cell-role classes, designed around one question: what does a cell contribute to a chunk? The two dominant classes are \emph{value} (observed data, 49\% of cells) and \emph{empty} (45\%). Cells whose content would only add noise to any chunk---placeholder text, decorative fragments---are labelled \emph{junk} (0.2\%) and excluded from chunking altogether. \emph{Aggregation} (0.9\%) separates formula-derived totals from raw values, since mistaking one for the other misleads numeric answers. The rest is structure: column and row headers, each at three nesting depths (4.0\% combined), plus sheet-level \emph{header}, \emph{metadata}, and \emph{comment} cells (0.4\%). Header depth matters because real sheets nest their headers; a single flat ``header'' class would collapse exactly the hierarchy that gives values their meaning. The distribution is highly imbalanced, which motivates the focal loss~\cite{lin2017focal} used in training.

Annotators also segment each sheet into logical tables (T0 for sheet-wide scope, T1+ for each distinct table), where a table is a contiguous region meant to be read as one set of records. This is what stops the chunker from ever mixing two unrelated tables in one chunk. Importantly, table membership is not a prediction target: the models are trained only on the 13 cell roles (the graph learners' auxiliary edge losses supervise cell adjacency and header--value links, never table ids), and the human table ids serve only for gold-role chunks and the oracle ceiling. At inference time, tables are recovered by a region detector that finds connected blocks of non-empty cells; we use Docling's Excel segmentation~\cite{docling2024}, but the pipeline is agnostic to the segmenter---any table-detection method can be plugged in, since the learned component is needed only for roles. Together, roles and table structure compactly encode the sheet's structure graph: cells are nodes, role labels type the nodes, and table membership plus header depth encode the header-owns-value edges that chunk assembly later follows.

\subsection{Learning cell roles: graph learners beat node classifiers}

To test whether recognizing cell roles improves spreadsheet RAG, we trained six role-annotation models belonging to two families. Node classifiers take table membership as given (from Docling's segmentation) and predict the 13 cell roles within tables: an MLP (three linear blocks, no message passing---the non-relational baseline), a GCN~\cite{kipf2017semi} (three GCNConv layers with LayerNorm residuals), and a GAT~\cite{velickovic2018gat} (three GATConv layers with a learned edge encoder). Graph learners additionally learn the sheet's structure itself---from value cells through multi-level row and column headers up to tables---by predicting cell-to-cell links alongside the roles: an AdjTransformer (a TransformerEncoder with four bilinear adjacency heads), a DualModalityGNN (separate content and format encoders fused over a learned $k$-NN graph), and a SpatialEdgeTransformer (spatial-bias attention with a typed edge scorer). All six consume the same 857-dimensional per-cell feature vector built from content, formatting, formula, and neighborhood signals; gold labels are never used as inputs. Full architecture details are in the supplementary material.

\begin{figure}[t]
  \centering
  \includegraphics[width=\linewidth]{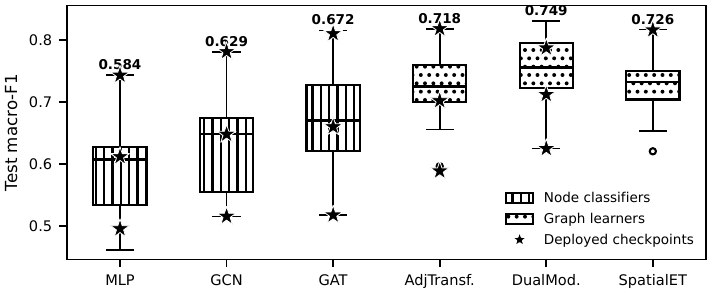}
  \caption{Test macro-F1 across 15 runs (5 folds $\times$ 3 seeds) per architecture: per-run pooled 13-class macro-F1, with means as data labels.}
  \label{fig:boxplot}
\end{figure}

Every architecture was trained under an identical grid on its training pool from \S 2.1: five-fold cross-validation at the sheet level, three seeds (42, 2137, 10042010), 100 epochs with AdamW and a cosine learning-rate schedule---90 runs in total. The loss is focal loss ($\gamma{=}2$) over the 13 classes with inverse-frequency weights, countering the heavy class imbalance; graph learners add a binary cross-entropy term on predicted adjacency, with gold header--value edges used only as auxiliary targets, never as inputs. We report per-run pooled macro-F1, i.e., the per-sheet confusion matrices of a run's test fold pooled into one matrix and macro-averaged over all 13 classes. As Figure~\ref{fig:boxplot} shows, every graph learner outperforms every node classifier (macro-F1 0.72--0.75 vs.\ 0.58--0.67, DualModalityGNN best at 0.75). Since the composition of tables varies from sheet to sheet, we hypothesize that learning the relationships between cells transfers across layouts better than classifying each cell in isolation, though we do not test this mechanism directly.

Table~\ref{tab:perclass} breaks the same scores down by class, with the best model per class in bold and fold-to-fold variance reported alongside the mean---for the rare classes (deeper header levels, comments, sheet headers, each under 1\% of cells) a handful of sheets can swing the score, so the variance is as informative as the mean. Rare classes are difficult for every model, which means the predicted structure our chunks are built from is never perfect and any downstream method must tolerate this noise; graph learners are nevertheless the best and most stable family, leading on the majority of classes with low to moderate variance.

\begin{table*}[t]
\centering
\caption{Per-class F1 (mean\,$\pm$\,SD over 15 runs = 5 folds $\times$ 3 seeds). Best mean per class in bold.}
\label{tab:perclass}
\small
\setlength{\tabcolsep}{4pt}
\begin{tabular}{@{}lr ccc ccc@{}}
\toprule
& & \multicolumn{3}{c}{\emph{node classifiers}} & \multicolumn{3}{c}{\emph{graph learners}}\\
\cmidrule(lr){3-5}\cmidrule(l){6-8}
Class & \%cells & MLP & GCN & GAT & DualMod. & AdjTr. & SpatET \\
\midrule
\texttt{empty}          & 47.5 & \ms{.970}{.024} & \ms{.966}{.033} & \ms{.975}{.031} & \msb{.979}{.008} & \ms{.975}{.013} & \ms{.973}{.012} \\
\texttt{value}          & 39.4 & \ms{.889}{.050} & \ms{.905}{.053} & \ms{.928}{.054} & \msb{.942}{.015} & \ms{.939}{.013} & \ms{.934}{.018} \\
\texttt{row\_header\_1} & 5.47 & \ms{.730}{.066} & \ms{.773}{.081} & \ms{.799}{.099} & \msb{.890}{.032} & \ms{.887}{.023} & \ms{.887}{.026} \\
\texttt{col\_header\_1} & 2.54 & \ms{.880}{.024} & \ms{.895}{.031} & \ms{.913}{.034} & \msb{.924}{.018} & \ms{.916}{.011} & \ms{.914}{.017} \\
\texttt{aggregation}    & 1.53 & \ms{.456}{.201} & \ms{.535}{.234} & \ms{.616}{.225} & \msb{.710}{.092} & \ms{.617}{.113} & \ms{.614}{.114} \\
\texttt{metadata}       & 1.46 & \ms{.589}{.157} & \ms{.656}{.149} & \ms{.735}{.162} & \msb{.849}{.027} & \ms{.833}{.030} & \ms{.827}{.035} \\
\texttt{row\_header\_2} & 0.61 & \ms{.295}{.131} & \ms{.325}{.148} & \ms{.396}{.187} & \msb{.574}{.115} & \ms{.486}{.149} & \ms{.511}{.142} \\
\texttt{junk}           & 0.57 & \ms{.419}{.134} & \ms{.459}{.119} & \ms{.465}{.129} & \msb{.480}{.156} & \ms{.455}{.153} & \ms{.453}{.137} \\
\texttt{header}         & 0.27 & \ms{.708}{.060} & \ms{.761}{.080} & \ms{.765}{.088} & \msb{.775}{.051} & \ms{.749}{.064} & \ms{.759}{.058} \\
\texttt{row\_header\_3} & 0.27 & \ms{.369}{.287} & \ms{.356}{.323} & \ms{.456}{.320} & \ms{.791}{.238} & \ms{.761}{.298} & \msb{.815}{.212} \\
\texttt{col\_header\_2} & 0.22 & \ms{.340}{.132} & \ms{.430}{.169} & \ms{.440}{.172} & \msb{.572}{.166} & \ms{.521}{.171} & \ms{.532}{.179} \\
\texttt{comment}        & 0.09 & \ms{.655}{.095} & \ms{.734}{.098} & \msb{.771}{.102} & \ms{.740}{.114} & \ms{.742}{.104} & \ms{.765}{.100} \\
\texttt{col\_header\_3} & 0.03 & \ms{.272}{.283} & \ms{.371}{.337} & \ms{.459}{.392} & \msb{.485}{.383} & \ms{.418}{.343} & \ms{.434}{.360} \\
\midrule
\textbf{Macro-F1}       &      & \ms{.584}{.075} & \ms{.629}{.081} & \ms{.672}{.091} & \msb{.749}{.057} & \ms{.718}{.064} & \ms{.726}{.049} \\
\bottomrule
\end{tabular}
\end{table*}

\subsection{Cell role annotation helps RAG}

We evaluated RAG quality end-to-end on the Q\&A dataset described above: 480 questions over 80 answer sheets, retrieved from the full 382-sheet corpus (answer sheets plus 302 distractors). For our approach, we deployed one trained checkpoint per architecture and built chunks from each checkpoint's predicted roles; for four of the six architectures the deployed checkpoint is the best cross-validation fold, and for GAT and DualModalityGNN a high-scoring fold (Figure~\ref{fig:boxplot} marks every deployed checkpoint as a star, including the additional low- and mid-quality checkpoints used in \S\ref{sec:generation}). We compared them against five baselines. Two are the most popular Python paths for ingesting spreadsheets in industry practice: BeautifulSoup, where the sheet is converted to an HTML table and parsed with bs4 (a general-purpose HTML parser widely used as the quick default for tabular ingestion), with each parsed row becoming a chunk; and Unstructured, a document-ingestion framework with native spreadsheet support. Three are academic approaches: STC~\cite{stc2026}, the strongest structure-aware baseline chunker; STC preceded by Docling's table split---so this variant runs STC within each detected table rather than on the whole sheet, using the same segmenter as our own pipeline and thereby isolating the effect of the roles from the effect of segmentation; and SpreadsheetLLM's SheetCompressor with Chain-of-Spreadsheet~\cite{dong2024spreadsheetllm}, modified to work over multiple sheets. We compared all methods on retrieval---recall@$k$, the share of questions for which a chunk containing the answer cells appears among the top $k$ retrieved results ($k{=}1$ and $5$)---and on answer quality, a final RAG rating from 1 to 5 given by a human judge to every generated answer.

\begin{table}[t]
\centering
\caption{End-to-end RAG results on 480 questions (human ratings, 1--5). Best per column in bold (oracle excluded).}
\label{tab:main}
\small
\begin{tabular}{@{}lccc@{}}
\toprule
\textbf{Method} & \textbf{Recall@1} & \textbf{Recall@5} & \textbf{Human} \\
\midrule
\multicolumn{4}{@{}l}{\emph{Baselines (no learned roles)}}\\
BeautifulSoup     & 0.219 & 0.400 & 1.91 \\
Unstructured      & 0.273 & 0.496 & 2.60 \\
SheetCompr.$+$CoS & 0.210 & 0.285 & 3.00 \\
STC $+$ Docling   & 0.325 & 0.492 & 3.34 \\
STC               & 0.356 & 0.515 & 3.43 \\
\midrule
\multicolumn{4}{@{}l}{\emph{Learned per-cell roles (ours; all six architectures)}}\\
MLP               & 0.467 & 0.613 & 3.52 \\
GCN               & 0.450 & 0.610 & 3.71 \\
DualModalityGNN   & \textbf{0.492} & 0.623 & 3.82 \\
GAT               & 0.465 & \textbf{0.640} & \textbf{3.88} \\
SpatialET         & 0.433 & 0.585 & 3.62 \\
AdjTransformer    & 0.458 & 0.625 & 3.75 \\
\midrule
Oracle (gold roles) & 0.502 & 0.608 & 4.01 \\
\bottomrule
\end{tabular}
\end{table}

Table~\ref{tab:main} shows the outcome. Every architecture scores at or above every baseline. The strongest checkpoint (GAT) reaches a human rating of 3.88 against STC's 3.43 (Wilcoxon $p = 1.5 \times 10^{-6}$) and dominates on retrieval as well (recall@1 0.465 vs.\ 0.356); DualModalityGNN, AdjTransformer, and GCN also beat STC significantly ($p \le 3{\times}10^{-3}$), SpatialET sits at the significance threshold ($p = 0.05$), and even the non-relational MLP matches it (3.52 vs.\ 3.43, $p = 0.43$). Additionally, to see how good RAG can get if the cell roles are known perfectly, we used the human role annotations of all 80 answer sheets to build gold-role chunks. The last row of Table~\ref{tab:main} shows this ceiling: 4.01---clearly above every learned checkpoint, yet still far from a perfect 5.0. So we know that interpreting spreadsheet structure through cell role annotation helps RAG; we do not yet know why, and we can also see that the approach has a limit even with perfect roles.

Role predictions by themselves do not dictate a chunk shape, and we had no a-priori reason to prefer one, so we treated the assembly geometry as an open choice and tested three over the same predicted roles (from the deployed GAT checkpoint of Table~\ref{tab:main}): \emph{Row}, which emits one chunk per record row and attaches to every value the full nesting path of its column and row headers; \emph{Flat}, which serializes each table region into a single chunk; and \emph{KG}, which decomposes tables into subject--predicate--object triples. In all three, chunks never cross the detected table boundaries, junk cells are dropped, and aggregation cells stay distinguishable from raw values. Each assembly was evaluated with the identical protocol: all 382 corpus sheets are chunked and indexed both densely (\texttt{intfloat/multilingual-e5-base}~\cite{wang2022e5}, 768-d) and lexically (BM25~\cite{robertson2009bm25}); for each question the top ten chunks are retrieved by reciprocal rank fusion~\cite{cormack2009rrf} ($k{=}60$), the top five go to the generator (\texttt{gemini-2.5-flash}, temperature 0), and every generated answer is rated 1--5 by a human judge.

\begin{table}[t]
\centering
\caption{Human rating (1--5) by chunk-assembly strategy and sheet structure ($n{=}480$).}
\label{tab:geometry}
\small
\begin{tabular}{@{}lccccc@{}}
\toprule
Sheet structure & $n$ & STC & Row & Flat & KG \\
\midrule
Nested (multi-level) headers & 81  & 3.12 & \textbf{4.14} & 3.35 & 2.43 \\
Matrix / cross-tab           & 200 & 3.71 & \textbf{4.13} & 3.81 & 3.20 \\
Multi-table                  & 86  & 4.03 & \textbf{4.23} & 3.67 & 1.66 \\
Simple flat table            & 99  & 3.82 & 3.81 & \textbf{3.84} & 2.11 \\
\midrule
All queries                  & 480 & 3.43 & \textbf{3.88} & 3.60 & 2.51 \\
\bottomrule
\end{tabular}
\end{table}

Table~\ref{tab:geometry} reports the outcome by table type: no single geometry is best for all table types. Row wins overall and on every structured type---nested headers, matrix, and multi-table sheets---but on simple flat tables the strategies converge to a tie, and the winning margin depends significantly on the table type (strategy$\times$type interaction, $p{=}1.0{\times}10^{-3}$). We therefore adopt Row as the default geometry throughout the paper while explicitly not claiming it is optimal: a single rigid assembly rule leaves value on the table, which is itself evidence that structure-conditioned, ideally learned, chunk assembly is the natural next step.

\subsection{Cell role annotation helps generation, not retrieval}
\label{sec:generation}

\begin{figure*}[t]
  \centering
  \includegraphics[width=0.92\textwidth]{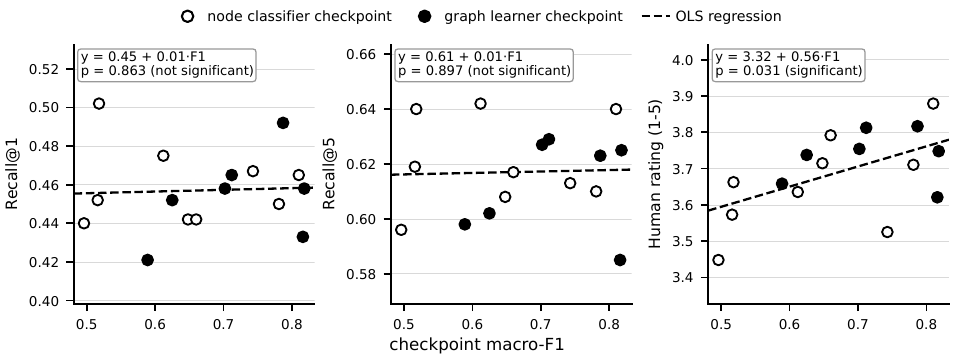}
  \caption{Sixteen deployed checkpoints, identical downstream pipeline: checkpoint macro-F1 against Recall@1, Recall@5, and the human answer rating, with OLS fits.}
  \label{fig:scatter}
\end{figure*}

Regardless of architecture, better role recognition makes for better RAG answers. The next logical step is to see why. To isolate the effect of role quality, we deploy sixteen checkpoints spanning a wide range of role quality (pooled macro-F1 0.50--0.82): low/mid/high triples within five architectures, plus a single checkpoint for the sixth (SpatialEdgeTransformer, whose memory limits made further runs unreliable). Each checkpoint runs the identical downstream pipeline---same Row chunk assembly, same corpus, same 480 questions---so the only thing that varies is how well the cell roles are recognized. For every checkpoint we measure retrieval (recall@1, recall@5) and human-rated answer quality (480 ratings per checkpoint), and fit an OLS regression of each metric on macro-F1 (Figure~\ref{fig:scatter}).

As the figure shows, the two effects split cleanly. Answer quality rises with role quality: the fit gains $+0.56$ rating points per unit of macro-F1 ($p{=}0.031$), and extrapolating it to perfect roles (F1${=}1.0$) predicts 3.9---consistent with the gold-role oracle, which actually attains 4.01: the ceiling sits on the same curve. Retrieval, in contrast, does not depend on role quality at all: both regression lines are flat (slopes ${\approx}\,0$, $p{=}0.86$ and $0.90$).

The reason is simple. A chunk matches a query lexically and semantically whether or not its cell roles were recognized correctly, so retrieval is blind to role quality. But for the LLM to interpret a retrieved value, the raw value alone is not enough---it also needs to know what the value means and where it sits: which column names it, which row it belongs to, which header hierarchy gives it units and scope. That context can only be packed into a chunk if the cell roles---and with them the table's structure---are recognized correctly. So good role recognition pays off when the answer is generated, not when the chunk is retrieved.

This is exactly what the opening example (Figure~\ref{fig:teaser}) illustrates. The three chunks shown there carry the same answer cell from the same nested-header sheet, and the sheet is findable for the retriever in every case---yet the chunks read very differently. Unstructured emits bare rows with the headers detached from their values; STC maps the values to generic first-row column labels that name nothing. Our Row chunk spells out the full context: the sheet title (the \emph{header} role marks sheet-level titles), the record's row header, and the complete column-header path above the value. Only the last one lets the generator tell which counterparty, which date, and which position a number belongs to without guessing---and that difference, invisible to retrieval, is precisely the gap the regression in Figure~\ref{fig:scatter} measures.

\subsection{Non-standard tables gain the most from predicted roles}

Knowing that roles help generation raises two follow-up questions: which roles carry the benefit, and on which sheets does it materialize? We answer the first with an ablation: starting from the gold-role chunks of the 80 answer sheets---the same human annotations that define the oracle ceiling, so no prediction error is involved---we remove one role class at a time by demoting every cell carrying it to plain \emph{value}, rebuild the chunks with the identical Row assembly, and rerun the identical retrieval and generation pipeline on all 480 questions. The \emph{judge loss} of a role is the drop in mean answer score relative to the unmodified gold chunks (baseline 3.66; panel scores sit systematically lower in level than the human columns of Table~\ref{tab:main}, but every judge loss is a panel-minus-panel difference, so the levels cancel), with significance from a Wilcoxon signed-rank test on the per-question paired differences. Because this is an auxiliary mechanism analysis with seven ablation arms ($7{\times}480$ answers), it is the one evaluation scored not by the human judge but by a panel of five versioned open-weight LLM judges using the identical 1--5 rubric; on answer sets carrying both panel and human ratings, the panel agrees with the human judge within one point on ${\geq}96\%$ of questions, so the substitution is safe for this auxiliary analysis.

\begin{table}[t]
\centering
\caption{Ablation on gold chunks: judge loss after removing one role class.}
\label{tab:ablation}
\small
\begin{tabular}{@{}lccl@{}}
\toprule
Role removed & judge loss & $p$ (Wilcoxon) & interpretation \\
\midrule
\texttt{col\_header\_1} & \textbf{1.40} & $2{\times}10^{-33}$ & names the values \\
\texttt{col\_header\_2} & 0.30 & $1{\times}10^{-8}$  & deeper column header \\
\texttt{row\_header\_2} & 0.22 & $9{\times}10^{-6}$  & deeper row header \\
\texttt{row\_header\_1} & 0.13 & $0.018$             & primary row header \\
\texttt{header}         & 0.12 & $0.023$             & sheet-level title \\
\texttt{aggregation}    & 0.06 & $0.038$             & formula-derived totals \\
\texttt{metadata}       & 0.06 & n.s.                & no measurable effect \\
\bottomrule
\end{tabular}
\end{table}

Table~\ref{tab:ablation} shows the result: chunk usefulness lives in the headers, and overwhelmingly in one of them. Demoting primary column headers (\texttt{col\_header\_1}) costs 1.40 rating points ($p{=}2{\times}10^{-33}$)---an order of magnitude more than any other role---because a value whose column is unnamed is just a number. The rest of the header hierarchy follows at a distance (\texttt{col\_header\_2} 0.30, \texttt{row\_header\_2} 0.22, \texttt{row\_header\_1} 0.13, sheet-level \texttt{header} 0.12), and it is exactly the part that only exists on sheets with nested structure: a simple table has no second header level to lose. Aggregation costs little (0.06) and metadata nothing measurable.

If the value of roles lies in recovering header hierarchy, the payoff should concentrate on sheets that have one---and it does. Table~\ref{tab:scope} breaks the human-rated advantage over the strongest baseline (STC) down by table type, alongside the gold-role ceiling. On nested-header sheets, perfect roles are worth $+1.15$ rating points over STC, and the best learned checkpoint already realizes $+1.01$ of that; matrix and multi-table sheets gain $+0.36$ at the ceiling. On simple flat tables the entire effect disappears: the ceiling shrinks to $+0.10$ and the best learned method to $-0.01$---a tie.

\begin{table}[t]
\centering
\caption{Human rating (1--5) by sheet structure: the strongest baseline (STC), our two best checkpoints, and the gold-role oracle.}
\label{tab:scope}
\footnotesize
\setlength{\tabcolsep}{3pt}
\begin{tabular}{@{}lcccccc@{}}
\toprule
Sheet structure & $n$ & STC & GAT & DualMod. & Oracle & $\Delta$ \\
\midrule
Nested headers     & 81  & 3.12 & 4.14 & 3.74 & \textbf{4.27} & $\mathbf{+1.15}$ \\
Matrix / cross-tab & 200 & 3.71 & \textbf{4.13} & 3.98 & 4.08 & $+0.36$ \\
Multi-table        & 86  & 4.03 & 4.23 & 4.12 & \textbf{4.40} & $+0.36$ \\
Simple flat table  & 99  & 3.82 & 3.81 & 3.91 & \textbf{3.92} & $+0.10$ \\
\bottomrule
\end{tabular}
\end{table}

This is no surprise. On a simple flat table, the first row already names every column, so a plain first-row heuristic recovers all the structure there is, and role annotation has nothing left to add. On non-standard layouts---nested headers, cross-tabs, multiple tables per sheet---that heuristic breaks down, and recovering the header hierarchy is worth up to a full rating point. Cell role annotation, in other words, is not a general-purpose booster: it is a targeted fix for exactly the tables that existing tools read wrong.

\subsection{Why our method wins---and why room for improvement remains}

We argue that the reason for this may be subtle: everyone who ever created a spreadsheet did it using their own preferences. Our semi-structured approach may not be enough for subtle differences between layouts and the infinite possibilities of table construction and interpretation. Even if we, as humans, read the table and assign roles, they may not exactly match the original author's idea.

We see that higher F1 reflects in better RAG quality, and that cell roles related to header order and relationships between values matter most for RAG quality. We see that the tables where we most outperform STC are the complicated, nested and messy ones. We think that pushing the approach further will require chunk assembly that adapts to each sheet's layout---up to rendering records as plain natural-language sentences---rather than committing to any fixed geometry in advance.

\section{Conclusion}

We showed that semantic cell-role annotation substantially improves spreadsheet RAG: chunks built from learned roles beat the strongest state-of-the-art chunker by $+0.45$ points on human-rated answer quality (3.88 vs.\ 3.43, $p{=}1.5{\times}10^{-6}$). The reason is not better retrievability but better interpretability of the chunks---each value arrives with the headers that explain it---and the benefit concentrates exactly where layouts are non-standard. We also measured the ceiling of the approach: even chunks built from perfect, human-annotated roles reach only 4.01 of 5. We argue this is because a fixed set of cell-role classes cannot capture the full variety of relationships that authors express in real-life spreadsheets, and because no single chunk geometry suits every table type---which together point toward learned, structure-conditioned rendering of 2D sheets into text as the way past the ceiling. To make the finding usable, we release our framework, which is modular in both the role annotator and the chunk geometry; in our experiments, the best configurations paired the GAT or DualModalityGNN annotator (statistically indistinguishable downstream) with row-based chunk assembly. The benchmark behind our evaluation---480 questions, gold role annotations, and roughly 18{,}000 human answer ratings---will be released separately as a dataset publication.

\bibliographystyle{ACM-Reference-Format}
\bibliography{references}

%% Appendix = former supplementary material, one-column for the wide tables/diagrams.
\clearpage
\onecolumn
\appendix
\input{appendix}

\end{document}

%% file: appendix.tex
%% Appendix of the arXiv preprint = the WSDM 2027 supplementary material, merged into
%% the main document (one-column). Sources: supplementary.tex + supp_arch.tex +
%% supp_classes.tex of the WSDM submission; numbered cross-references replaced by \ref.

\input{supp_arch}

\section{Training hyperparameters}

All 90 runs (6 architectures $\times$ 5 folds $\times$ 3 seeds) share one grid:

\begin{itemize}
  \item Cross-validation: 5 folds, split at the sheet level; seeds 42, 2137, 10042010.
  \item Epochs: 100; optimizer AdamW; cosine learning-rate schedule.
  \item Sheets capped at 800 nodes during training.
  \item Loss: focal loss ($\gamma{=}2$) over the 13 classes with inverse-frequency $\alpha$ weights. Graph learners add binary cross-entropy on predicted adjacency with warmup scaling; gold header--value hierarchy edges serve only as auxiliary loss targets and are excluded from the convolution graph.
  \item Hidden dimension: 128 (node classifiers), 256 (graph learners); dropout in the input encoder and backbone blocks.
  \item Reported metric: per-run pooled macro-F1 --- the per-sheet 13-class confusion matrices of the run's test fold are pooled into a single matrix, per-class F1 is computed over classes with gold support, and averaged.
\end{itemize}

\begin{figure}[H]
  \centering
  \includegraphics[width=0.8\linewidth]{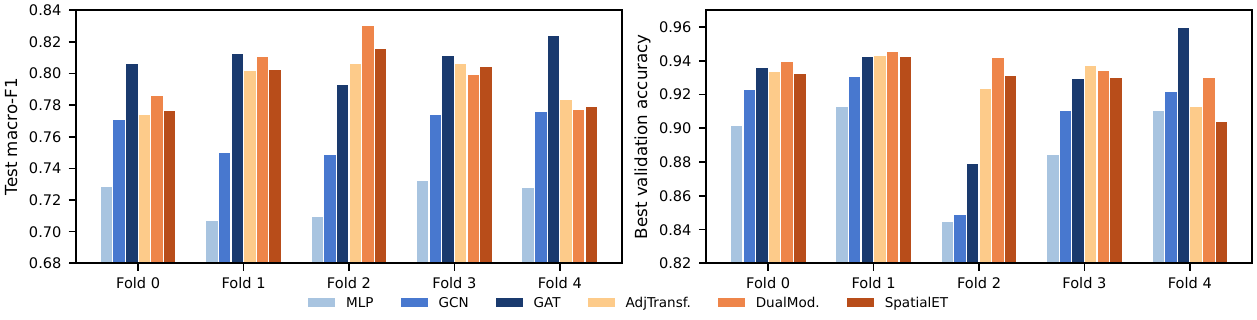}
  \caption{Per-fold test macro-F1 (left) and best validation accuracy (right), averaged over 3 seeds. Blue = node classifiers, orange = graph learners.}
  \label{fig:foldvar}
\end{figure}

Figure~\ref{fig:foldvar} disaggregates performance by fold: fold~2 is consistently the hardest, and the architecture ranking is stable across folds.

\input{supp_classes}

\section{Annotation protocol}

Every sheet used for training or evaluation was structured by two human labelers working in native Excel, following a published bilingual labelling guide; the workflow includes peer review and a self-review pass. Annotators could mark a tab ``skipped'' when they could not interpret how to read it coherently; skipped tabs (${\approx}24\%$ of decided tabs) are excluded from training, evaluation, and RAG indexing. Labels attach per-sheet table ids (T0 = sheet-wide scope; T1+ = distinct logical tables) and header depth levels for row and column headers. Table membership is never used as a model input or prediction target; it serves only for building gold-role chunks and the oracle ceiling.

\section{Deployed checkpoints (relationship study)}

Table~\ref{tab:supp_ckpt} lists all sixteen deployed checkpoints used in the role-quality regression (Section~\ref{sec:generation}), with their pooled macro-F1 and downstream results under the identical Row pipeline. MLP, GCN, GAT, AdjTransformer, and DualModalityGNN contribute low/mid/high triples; SpatialEdgeTransformer contributes a single checkpoint (its $O(N^2)$ edge scorer required a stricter node cap: 138/382 corpus sheets fell back to naive chunks vs.\ 33/382 for all other checkpoints, so its scores are partly fallback-driven).

\begin{table}[H]
\centering
\caption{The sixteen deployed checkpoints, sorted by pooled macro-F1. Human = mean human rating over 480 questions.}
\label{tab:supp_ckpt}
\small
\begin{tabular}{@{}llcccc@{}}
\toprule
Checkpoint & Family & macro-F1 & Recall@1 & Recall@5 & Human \\
\midrule
MLP-low      & node  & 0.496 & 0.440 & 0.596 & 3.45 \\
GCN-low      & node  & 0.516 & 0.452 & 0.619 & 3.57 \\
GAT-low      & node  & 0.518 & 0.502 & 0.640 & 3.66 \\
AdjTr-low    & graph & 0.589 & 0.421 & 0.598 & 3.66 \\
MLP-mid      & node  & 0.612 & 0.475 & 0.642 & 3.64 \\
Dual-low     & graph & 0.625 & 0.452 & 0.602 & 3.74 \\
GCN-mid      & node  & 0.648 & 0.442 & 0.608 & 3.71 \\
GAT-mid      & node  & 0.660 & 0.442 & 0.617 & 3.79 \\
AdjTr-mid    & graph & 0.702 & 0.458 & 0.627 & 3.75 \\
Dual-mid     & graph & 0.712 & 0.465 & 0.629 & 3.81 \\
MLP-high     & node  & 0.743 & 0.467 & 0.613 & 3.52 \\
GCN-high     & node  & 0.781 & 0.450 & 0.610 & 3.71 \\
Dual-high    & graph & 0.787 & 0.492 & 0.623 & 3.82 \\
GAT-high     & node  & 0.810 & 0.465 & 0.640 & 3.88 \\
SpatialET    & graph & 0.816 & 0.433 & 0.585 & 3.62 \\
AdjTr-high   & graph & 0.818 & 0.458 & 0.625 & 3.75 \\
\bottomrule
\end{tabular}
\end{table}

\section{Prompts}

\subsection{Answer generation prompt}

The generator (\texttt{gemini-2.5-flash}, thinking disabled, temperature 0) receives the top five retrieved chunks joined with \texttt{---} separators:

\begin{Verbatim}[frame=single, fontsize=\small]
You are answering questions about data in spreadsheets.
Given the following spreadsheet excerpts, answer the question.
Include the specific values and numbers from the data that
support your answer.
If you cannot find the answer in the provided excerpts, say
"NOT FOUND".

Spreadsheet excerpts:
{chunks}

Question: {question}

Answer:
\end{Verbatim}

\subsection{Judge prompt (role-ablation panel)}

The role ablation (Table~\ref{tab:ablation}) is the one evaluation scored by a panel of five open-weight LLM judges instead of the human judge, for three reasons. First, scale: the ablation adds $7{\times}480 = 3{,}360$ generated answers for a single auxiliary table --- an extra rating round of about a fifth of the entire human-rated volume, disproportionate to the weight of the result. Second, the analysis is purely \emph{relative}: what matters is the difference between ablation arms on identical questions and identical gold chunks, so a consistent, repeatable judge suffices and no absolute calibration is needed --- and the ablation baseline (unmodified gold chunks) is scored by the same panel, so every reported judge loss is a panel-minus-panel difference, free of any human-vs-panel offset. Third, the substitution is validated: on answer sets carrying both panel and human ratings, the panel agrees with the human judge within one point on ${\geq}96\%$ of questions and runs ${\sim}0.1$ points lower on average. Each judge receives the same rubric that the human judge uses:

\begin{Verbatim}[frame=single, fontsize=\small]
You are evaluating whether a generated answer correctly answers
the same question as an expected answer.
Both answers refer to data extracted from a spreadsheet.

Expected answer: {expected}
Generated answer: {generated}

Rate the generated answer on a scale of 1-5:
1 = Completely wrong, contradicts the expected answer, or
    unrelated
2 = Partially related but key facts are wrong or misleading
3 = Answers a different aspect of the question, or gives only
    tangentially related information
4 = Correctly answers the core question but with less detail or
    supporting data than the expected answer
5 = Correctly answers the question with equivalent or sufficient
    detail

Important: if the question is yes/no or asks "which one", and the
generated answer gives the correct yes/no/choice, that is at
least a 4 even if it omits supporting numbers. The core answer
matters most.

Respond with ONLY a single integer (1-5).
\end{Verbatim}

\section{Human evaluation protocol}

Every generated answer in every reported human column --- ours, every baseline's, every geometry variant's, and the oracle's --- was rated on the same 1--5 rubric shown above by a human judge, blind to which method produced the chunks. The only exception is the seven-arm role ablation (Table~\ref{tab:ablation}), which is scored by a panel of five versioned open-weight LLM judges (DeepSeek-v4-flash, Llama-3.3-70B, Qwen3.7-flash, Gemma-3-27B, Mistral-Small-24B) using the identical rubric; on answer sets carrying both panel and human ratings, the panel agrees with the human judge within one point on ${\geq}96\%$ of questions and runs ${\sim}0.1$ points lower on average, so the substitution is safe for this auxiliary analysis.

%% file: supp_arch.tex
\section{Architecture details}
\label{app:arch_details}

All models share a common interface: \texttt{forward} produces node embeddings $h$, \texttt{classify\_nodes} maps $h$ to 13-class logits, and \texttt{predict\_edges} produces hierarchy and table-boundary edge logits.
Table~\ref{tab:arch_details} summarises backbone layers, total learned depth (including encoder, classification head, edge heads, and auxiliary modules), and parameter counts.

\begin{table}[H]
\centering
\caption{Architecture specifications. Hidden dim is 128 for node classifiers, 256 for graph learners. All models receive the same 857-d input feature vector. ``Depth'' counts backbone layers plus input encoder, classification head, edge heads, and (for graph learners) adjacency heads and optional refinement layers.}
\label{tab:arch_details}
\footnotesize
\setlength{\tabcolsep}{3pt}
\begin{tabular}{@{}llcccp{5cm}@{}}
\toprule
\textbf{Family} & \textbf{Architecture} & \textbf{Backbone} & \textbf{Depth} & \textbf{Params} & \textbf{Key components} \\
\midrule
\multirow{3}{*}{\shortstack[l]{Node\\classifiers}}
& MLP & 3 & 7 & 718\,K & Linear $\to$ LayerNorm $\to$ ReLU $\to$ Dropout; no message passing \\
& GCN & 3 & 8 & 784\,K & GCNConv with symmetric normalisation, residual + LayerNorm per layer \\
& GAT & 3 & 9 & 2.18\,M & GATConv (4 heads, averaged), learned edge encoder (23-d $\to$ hidden), residual + LayerNorm \\
\midrule
\multirow{3}{*}{\shortstack[l]{Graph\\learners}}
& AdjTransformer & 4 & ${\sim}$15 & 3.57\,M & Row/col pos.\ embeddings, TransformerEncoder (4 heads, GELU), 4 bilinear adjacency heads, optional StructureRefinementGNN \\
& DualModalityGNN & 2 & ${\sim}$17 & 1.93\,M & Dual content/format encoders, bilinear fusion, learned $k$-NN graph, graph reasoning layers with edge MLPs \\
& SpatialEdgeTransformer & 4 & ${\sim}$14 & 3.62\,M & Row/col pos.\ embeddings, TransformerEncoder (4 heads), spatial-bias attention (7-d pair features), typed MLP edge scorer \\
\bottomrule
\end{tabular}
\end{table}

The six architectures split into two families that differ in \emph{where the graph comes from}.
Node classifiers consume a fixed, hand-engineered adjacency (dashed gray in the diagrams); graph learners predict the graph topology via learned adjacency heads (yellow blocks) and optionally refine it through a second-pass GNN (cyan).
Figure~\ref{fig:families_comparison} contrasts the two patterns side by side; the same colour coding is used in all six per-architecture diagrams that follow.

\begin{figure}[H]
\centering
\begin{subfigure}[t]{0.44\textwidth}
\centering
\begin{tikzpicture}[
  block/.style={draw, rounded corners, minimum width=2.2cm, minimum height=0.6cm,
                align=center, font=\footnotesize},
  arrow/.style={-{Stealth[length=2mm]}, thick},
  node distance=0.5cm,
  scale=0.92, every node/.style={scale=0.92}
]
\node[block, fill=blue!8] (input) {857-d input};
\node[block, fill=green!10, below=of input] (enc) {Encoder};
\node[block, fill=orange!12, below=of enc] (bb) {Backbone\\{\scriptsize conv / linear}};
\node[block, fill=gray!8, dashed, left=0.8cm of bb, minimum width=1.7cm] (adj) {Engineered\\adjacency\\{\scriptsize (fixed)}};
\node[block, fill=red!8, below=0.5cm of bb] (head) {Node head\\{\scriptsize $\to$ 13 logits}};
\node[block, fill=purple!8, right=0.8cm of head] (edge) {Edge heads};
\draw[arrow] (input) -- (enc);
\draw[arrow] (enc) -- (bb);
\draw[-{Stealth[length=2mm]}, thick, dashed] (adj) -- (bb);
\draw[arrow] (bb) -- (head);
\draw[arrow] (bb.east) -- ++(0.4,0) |- (edge.west);
\end{tikzpicture}
\subcaption{Node classifier pattern}
\label{fig:family_nc}
\end{subfigure}
\hfill
\begin{subfigure}[t]{0.53\textwidth}
\centering
\begin{tikzpicture}[
  block/.style={draw, rounded corners, minimum width=2.2cm, minimum height=0.6cm,
                align=center, font=\footnotesize},
  arrow/.style={-{Stealth[length=2mm]}, thick},
  node distance=0.5cm,
  scale=0.92, every node/.style={scale=0.92}
]
\node[block, fill=blue!8] (input) {857-d input};
\node[block, fill=green!10, below=of input] (proj) {Projection};
\node[block, fill=orange!12, below=of proj] (bb) {Backbone\\{\scriptsize transformer}};
\node[block, fill=yellow!15, right=1.0cm of bb] (adjh) {Adjacency\\heads\\{\scriptsize (learned)}};
\node[block, fill=cyan!10, below=0.5cm of bb] (refine) {Refinement\\GNN\\{\scriptsize (optional)}};
\node[block, fill=red!8, below=0.5cm of refine] (head) {Node head\\{\scriptsize $\to$ 13 logits}};
\node[block, fill=purple!8, right=1.0cm of head] (edge) {Edge heads};
\draw[arrow] (input) -- (proj);
\draw[arrow] (proj) -- (bb);
\draw[arrow] (bb.east) -- ++(0.5,0) |- (adjh.west);
\draw[-{Stealth[length=2mm]}, line width=1.2pt] (adjh.south) -- ++(0,-0.25) -| (refine.east) node[pos=0.65, below, font=\scriptsize\bfseries] {learned adjacency};
\draw[arrow] (bb) -- (refine);
\draw[arrow] (refine) -- (head);
\draw[arrow] (refine.east) -- ++(0.5,0) |- (edge.west);
\end{tikzpicture}
\subcaption{Graph learner pattern}
\label{fig:family_gl}
\end{subfigure}
\caption{The defining architectural difference. \textbf{Left:} node classifiers receive a fixed, hand-engineered adjacency graph (dashed gray); the model only learns node representations over this static wiring. \textbf{Right:} graph learners predict the graph topology via learned adjacency heads (yellow); predicted edges feed an optional refinement GNN (cyan) before classification. The bold ``learned adjacency'' arrow is the feedback path absent from node classifiers.}
\label{fig:families_comparison}
\end{figure}

\paragraph{Colour key.}
All architecture diagrams use the same block colours:
\textcolor{blue!50}{\rule{0.8em}{0.8em}}~input features,
\textcolor{green!40}{\rule{0.8em}{0.8em}}~encoder / projection,
\textcolor{orange!40}{\rule{0.8em}{0.8em}}~backbone (conv or transformer),
{\setlength{\fboxsep}{0pt}\fbox{\textcolor{gray!20}{\rule{0.8em}{0.8em}}}}~fixed engineered graph (node classifiers only; dashed border),
\textcolor{yellow!50}{\rule{0.8em}{0.8em}}~\textbf{learned adjacency heads} (graph learners only),
\textcolor{cyan!30}{\rule{0.8em}{0.8em}}~optional refinement GNN,
\textcolor{red!25}{\rule{0.8em}{0.8em}}~classification head,
\textcolor{purple!25}{\rule{0.8em}{0.8em}}~edge prediction heads.

%% ---------- Node classifier family ----------
\subsection{Node classifiers}

Node classifiers operate on the fixed engineered graph (grid neighbours, row/column strips, sheet-to-table bridges) and do not modify graph topology.
All three share a common structure: an input encoder (Linear + LayerNorm + ReLU + Dropout) projects the 857-d feature vector to the hidden dimension, a backbone processes node representations, and a 2-layer MLP classification head maps to 13 role logits.
Two additional 2-layer MLP heads predict hierarchy and table-boundary edges.
Figures~\ref{fig:arch_mlp}--\ref{fig:arch_gat} show each architecture individually.

\paragraph{MLP (Figure~\ref{fig:arch_mlp}).}
The simplest baseline: three fully-connected blocks (Linear $\to$ LayerNorm $\to$ ReLU $\to$ Dropout) classify each cell independently from its feature vector, with no message passing.
Despite ignoring graph structure entirely, MLP provides a strong feature-only baseline that isolates the contribution of the 857-d cell representation.
As the diagram shows, the backbone is purely feed-forward --- no neighbour information enters the computation at any stage.

\begin{figure}[H]
\centering
\begin{tikzpicture}[
  block/.style={draw, rounded corners, minimum width=3.2cm, minimum height=0.7cm,
                align=center, font=\small},
  arrow/.style={-{Stealth[length=2.5mm]}, thick},
  node distance=0.6cm
]
\node[block, fill=blue!8] (input) {857-d input};
\node[block, fill=green!10, below=of input] (enc) {Input encoder\\{\scriptsize Linear + LN + ReLU + Drop}};
\node[block, fill=orange!12, below=of enc] (fc) {Linear blocks $\times 3$\\{\scriptsize LN + ReLU + Drop each}\\{\scriptsize \textbf{no message passing}}};
\node[block, fill=red!8, below=0.6cm of fc] (head) {Node head\\{\scriptsize 2-layer MLP $\to$ 13 logits}};
\node[block, fill=purple!8, right=1.8cm of head] (edge) {Edge heads $\times 2$\\{\scriptsize hierarchy + table bdy}};
\node[block, fill=gray!8, dashed, left=1.5cm of fc, minimum width=2cm] (adj) {No adjacency\\{\scriptsize (no graph)}};
\draw[arrow] (input) -- (enc);
\draw[arrow] (enc) -- (fc);
\draw[arrow] (fc) -- (head);
\draw[arrow] (fc.east) -- ++(0.9,0) |- (edge.west);
\draw[dashed, thick, gray!50] (adj) -- (fc) node[midway, above, font=\tiny, text=gray!70] {$\times$};
\end{tikzpicture}
\caption{\textbf{MLP} (718\,K params). Each cell is classified independently from its 857-d feature vector; three linear blocks replace convolution. The dashed gray box with ``$\times$'' emphasises that no graph structure is used --- this is a non-relational baseline.}
\label{fig:arch_mlp}
\end{figure}

\paragraph{GCN (Figure~\ref{fig:arch_gcn}).}
Three GCNConv layers with symmetric normalisation propagate information along the engineered graph.
Each layer applies a residual connection followed by LayerNorm, enabling gradient flow through the message-passing stack.
The key difference from MLP is that GCN aggregates features from spatial neighbours via the fixed adjacency, so each cell's representation reflects its local context.

\begin{figure}[H]
\centering
\begin{tikzpicture}[
  block/.style={draw, rounded corners, minimum width=3.2cm, minimum height=0.7cm,
                align=center, font=\small},
  arrow/.style={-{Stealth[length=2.5mm]}, thick},
  node distance=0.6cm
]
\node[block, fill=blue!8] (input) {857-d input};
\node[block, fill=green!10, below=of input] (enc) {Input encoder\\{\scriptsize Linear + LN + ReLU + Drop}};
\node[block, fill=orange!12, below=of enc] (gcn) {GCNConv $\times 3$\\{\scriptsize symmetric norm}\\{\scriptsize residual + LN each}};
\node[block, fill=red!8, below=0.6cm of gcn] (head) {Node head\\{\scriptsize 2-layer MLP $\to$ 13 logits}};
\node[block, fill=purple!8, right=1.8cm of head] (edge) {Edge heads $\times 2$\\{\scriptsize hierarchy + table bdy}};
\node[block, fill=gray!8, dashed, left=1.5cm of gcn, minimum width=2cm] (adj) {Engineered\\adjacency\\{\scriptsize (fixed)}};
\draw[arrow] (input) -- (enc);
\draw[arrow] (enc) -- (gcn);
\draw[arrow] (gcn) -- (head);
\draw[arrow] (gcn.east) -- ++(0.9,0) |- (edge.west);
\draw[-{Stealth[length=2.5mm]}, thick, dashed] (adj) -- (gcn);
\end{tikzpicture}
\caption{\textbf{GCN} (784\,K params). Three GCNConv layers propagate features over the fixed engineered adjacency (grid, row/col strips, sheet-to-table bridges). The dashed box indicates the static graph that is not learned.}
\label{fig:arch_gcn}
\end{figure}

\paragraph{GAT (Figure~\ref{fig:arch_gat}).}
Three GATConv layers with 4 attention heads (averaged) learn to weight neighbour messages.
A learned edge encoder maps the 23-d edge feature vector (type, relative position) to the hidden dimension, providing edge-type-aware attention.
Each layer includes residual connections and LayerNorm.
GAT is the largest node classifier (2.18\,M parameters) due to the multi-head attention and edge encoding parameters.
The dedicated edge encoder, shown in the diagram as a separate block feeding into each GATConv layer, is the distinguishing feature over GCN.

\begin{figure}[H]
\centering
\begin{tikzpicture}[
  block/.style={draw, rounded corners, minimum width=3.2cm, minimum height=0.7cm,
                align=center, font=\small},
  arrow/.style={-{Stealth[length=2.5mm]}, thick},
  node distance=0.6cm
]
\node[block, fill=blue!8] (input) {857-d input};
\node[block, fill=green!10, below=of input] (enc) {Input encoder\\{\scriptsize Linear + LN + ReLU + Drop}};
\node[block, fill=orange!12, below=of enc] (gat) {GATConv $\times 3$\\{\scriptsize 4 heads, averaged}\\{\scriptsize residual + LN each}};
\node[block, fill=orange!6, right=1.8cm of gat] (edgeenc) {Edge encoder\\{\scriptsize 23-d $\to$ hidden}};
\node[block, fill=red!8, below=0.6cm of gat] (head) {Node head\\{\scriptsize 2-layer MLP $\to$ 13 logits}};
\node[block, fill=purple!8, right=1.8cm of head] (edge) {Edge heads $\times 2$\\{\scriptsize hierarchy + table bdy}};
\node[block, fill=gray!8, dashed, left=1.5cm of gat, minimum width=2cm] (adj) {Engineered\\adjacency\\{\scriptsize (fixed)}};
\draw[arrow] (input) -- (enc);
\draw[arrow] (enc) -- (gat);
\draw[arrow] (gat) -- (head);
\draw[arrow] (gat.east) -- ++(0.9,0) |- (edge.west);
\draw[arrow] (edgeenc) -- (gat);
\draw[-{Stealth[length=2.5mm]}, thick, dashed] (adj) -- (gat);
\end{tikzpicture}
\caption{\textbf{GAT} (2.18\,M params). Three GATConv layers with 4 attention heads learn to weight neighbour messages over the fixed adjacency. A learned edge encoder maps each edge's 23-d feature vector (type identifier + relative position) into the hidden dimension, enabling edge-type-aware attention --- the key addition over GCN.}
\label{fig:arch_gat}
\end{figure}

%% ---------- Graph learner family ----------
\subsection{Graph learners}

Graph learners additionally predict soft adjacency matrices and receive edge-level supervision for four edge types (table membership, hierarchy, same-row, same-column).
They produce $N{\times}N$ logit matrices via learned heads, optionally refine predictions through a StructureRefinementGNN, and jointly optimise node classification and edge prediction losses.
Figures~\ref{fig:arch_adjtransf}--\ref{fig:arch_spatialet} show each architecture; the same colour coding as the node classifiers applies, with two additional colours: yellow for adjacency prediction heads and cyan for the optional refinement GNN.

\paragraph{AdjTransformer (Figure~\ref{fig:arch_adjtransf}).}
An input projection maps the 857-d features to the hidden dimension, augmented with learned row and column positional embeddings.
Four TransformerEncoder layers (4 heads, GELU activation) produce contextualised node representations.
Four bilinear adjacency heads compute $\sigma(Q_k K_k^\top)$ matrices, one per edge type.
When \texttt{struct\_refine\_active} is set, predicted edges above a learned threshold are fed into a 2-layer StructureRefinementGNN for second-pass message passing before the classification head.
The bilinear heads (right branch in the diagram) are the distinguishing mechanism: each head learns a separate query--key space for one edge type.

\begin{figure}[H]
\centering
\begin{tikzpicture}[
  block/.style={draw, rounded corners, minimum width=3.2cm, minimum height=0.7cm,
                align=center, font=\small},
  arrow/.style={-{Stealth[length=2.5mm]}, thick},
  node distance=0.6cm
]
\node[block, fill=blue!8] (input) {857-d input};
\node[block, fill=green!10, below=of input] (proj) {Input projection\\{\scriptsize + row/col pos.\ embed}};
\node[block, fill=orange!12, below=of proj] (transf) {TransformerEncoder $\times 4$\\{\scriptsize 4 heads, GELU}};
\node[block, fill=yellow!15, right=2cm of transf] (adj) {Bilinear adj.\ heads $\times 4$\\{\scriptsize $\sigma(Q_k K_k^\top)$ per edge type}};
\node[block, fill=cyan!10, below=0.6cm of transf] (refine) {StructureRefinementGNN\\{\scriptsize (optional, 2 layers)}};
\node[block, fill=red!8, below=0.6cm of refine] (head) {Node head\\{\scriptsize 2-layer MLP $\to$ 13 logits}};
\node[block, fill=purple!8, right=2cm of head] (edge) {Edge heads $\times 2$\\{\scriptsize hierarchy + table bdy}};
\draw[arrow] (input) -- (proj);
\draw[arrow] (proj) -- (transf);
\draw[arrow] (transf) -- (refine);
\draw[arrow] (transf.east) -- ++(1.0,0) |- (adj.west);
\draw[-{Stealth[length=2.5mm]}, line width=1.2pt] (adj.south) -- ++(0,-0.35) -| (refine.east) node[pos=0.25, above, font=\scriptsize\bfseries] {learned adjacency};
\draw[arrow] (refine) -- (head);
\draw[arrow] (refine.east) -- ++(1.0,0) |- (edge.west);
\end{tikzpicture}
\caption{\textbf{AdjTransformer} (3.57\,M params). A 4-layer TransformerEncoder produces node representations; four bilinear adjacency heads predict edge types via $\sigma(QK^\top)$. The bold ``learned adjacency'' arrow feeds predicted edges into an optional 2-layer StructureRefinementGNN before the classification head --- no fixed graph is provided.}
\label{fig:arch_adjtransf}
\end{figure}

\paragraph{DualModalityGNN (Figure~\ref{fig:arch_dualmgnn}).}
Content statistics and formatting features are encoded through separate 2-layer pathways and merged via a bilinear fusion block.
A learned $k$-nearest-neighbour graph is constructed from fused embeddings.
Two graph reasoning layers (each containing an edge MLP, attention-weighted aggregation, and node update) propagate information over the combined topology.
Four adjacency heads (same bilinear $\sigma(QK^\top)$ pattern) predict edge types.
DualModalityGNN has the most complex input processing (${\sim}$17 total learned layers) but the fewest parameters among graph learners (1.93\,M) because it uses a smaller hidden dimension for the dual encoders.
The dual-path input stage, shown as two parallel encoder blocks in the diagram, is the defining design choice: it forces the model to learn modality-specific representations before fusion.

\begin{figure}[H]
\centering
\begin{tikzpicture}[
  block/.style={draw, rounded corners, minimum width=3.0cm, minimum height=0.7cm,
                align=center, font=\small},
  arrow/.style={-{Stealth[length=2.5mm]}, thick},
  node distance=0.6cm
]
\node[block, fill=blue!8, minimum width=5cm] (input) {857-d input};
\node[block, fill=green!12, below left=0.7cm and 0.3cm of input] (content) {Content encoder\\{\scriptsize 2-layer MLP}};
\node[block, fill=green!18, below right=0.7cm and 0.3cm of input] (format) {Format encoder\\{\scriptsize 2-layer MLP}};
\node[block, fill=green!10, below=1.8cm of input] (fusion) {Bilinear fusion};
\node[block, fill=orange!12, below=of fusion] (knn) {Learned $k$-NN graph\\{\scriptsize + 2 graph reasoning layers}\\{\scriptsize edge MLP + attn agg.}};
\node[block, fill=yellow!15, right=2cm of knn] (adj) {Adjacency heads $\times 4$\\{\scriptsize $\sigma(QK^\top)$ per edge type}};
\node[block, fill=red!8, below=0.6cm of knn] (head) {Node head\\{\scriptsize 2-layer MLP $\to$ 13 logits}};
\node[block, fill=purple!8, right=2cm of head] (edge) {Edge heads $\times 2$\\{\scriptsize hierarchy + table bdy}};
\draw[arrow] (input.south) -- ++(-1.2,-0.3) -- (content.north);
\draw[arrow] (input.south) -- ++(1.2,-0.3) -- (format.north);
\draw[arrow] (content.south) -- ++(0.0,-0.15) -| (fusion.north west);
\draw[arrow] (format.south) -- ++(0.0,-0.15) -| (fusion.north east);
\draw[arrow] (fusion) -- (knn);
\draw[arrow] (knn) -- (head);
\draw[arrow] (knn.east) -- ++(1.0,0) |- (adj.west);
\draw[-{Stealth[length=2.5mm]}, line width=1.2pt] (adj.south) -- ++(0,-0.6) -| (knn.south east) node[pos=0.2, right, font=\scriptsize\bfseries] {learned adjacency};
\draw[arrow] (knn.east) -- ++(1.0,0) |- (edge.west);
\end{tikzpicture}
\caption{\textbf{DualModalityGNN} (1.93\,M params). Content and format features are encoded through separate pathways, fused via a bilinear block, and processed by graph reasoning layers over a learned $k$-NN topology. Adjacency heads predict edge types and feed \textbf{learned adjacency} back into the graph reasoning layers --- no fixed graph is provided. The dual-encoder input stage is the distinguishing design.}
\label{fig:arch_dualmgnn}
\end{figure}

\paragraph{SpatialEdgeTransformer (Figure~\ref{fig:arch_spatialet}).}
An input projection plus learned row/column positional embeddings feed into four TransformerEncoder layers (4 heads).
For edge prediction, 7-dimensional spatial pair features (relative row, relative column, Manhattan distance, Chebyshev distance, same-row flag, same-column flag, log-area ratio) are computed for all candidate pairs within a configurable grid radius (default 50).
A typed pairwise MLP scores each candidate edge, replacing the bilinear heads used by AdjTransformer.
An optional 2-layer StructureRefinementGNN post-processes the predicted graph before the classification head.
The 7-d spatial pair features, shown as a dedicated input to the MLP edge scorer in the diagram, make this architecture the most spatially explicit among graph learners.

\begin{figure}[H]
\centering
\begin{tikzpicture}[
  block/.style={draw, rounded corners, minimum width=3.2cm, minimum height=0.7cm,
                align=center, font=\small},
  arrow/.style={-{Stealth[length=2.5mm]}, thick},
  node distance=0.6cm
]
\node[block, fill=blue!8] (input) {857-d input};
\node[block, fill=green!10, below=of input] (proj) {Input projection\\{\scriptsize + row/col pos.\ embed}};
\node[block, fill=orange!12, below=of proj] (transf) {TransformerEncoder $\times 4$\\{\scriptsize 4 heads, spatial-bias attn}};
\node[block, fill=yellow!15, right=2cm of transf] (mlpedge) {Typed MLP edge scorer\\{\scriptsize per edge type}};
\node[block, fill=yellow!8, above=0.3cm of mlpedge] (spatial) {7-d spatial pair features\\{\scriptsize $\Delta$row, $\Delta$col, Manhattan,}\\{\scriptsize Chebyshev, flags, log-area}};
\node[block, fill=cyan!10, below=0.6cm of transf] (refine) {StructureRefinementGNN\\{\scriptsize (optional, 2 layers)}};
\node[block, fill=red!8, below=0.6cm of refine] (head) {Node head\\{\scriptsize 2-layer MLP $\to$ 13 logits}};
\node[block, fill=purple!8, right=2cm of head] (edge) {Edge heads $\times 2$\\{\scriptsize hierarchy + table bdy}};
\draw[arrow] (input) -- (proj);
\draw[arrow] (proj) -- (transf);
\draw[arrow] (transf) -- (refine);
\draw[arrow] (transf.east) -- ++(1.0,0) |- (mlpedge.west);
\draw[arrow] (spatial) -- (mlpedge);
\draw[-{Stealth[length=2.5mm]}, line width=1.2pt] (mlpedge.south) -- ++(0,-0.35) -| (refine.east) node[pos=0.25, above, font=\scriptsize\bfseries] {learned adjacency};
\draw[arrow] (refine) -- (head);
\draw[arrow] (refine.east) -- ++(1.0,0) |- (edge.west);
\end{tikzpicture}
\caption{\textbf{SpatialEdgeTransformer} (3.62\,M params). Shares the Transformer backbone with AdjTransformer but replaces bilinear adjacency heads with a typed MLP edge scorer that consumes 7-d spatial pair features (relative position, distances, flags). The bold ``learned adjacency'' arrow feeds predicted edges into the refinement GNN --- no fixed graph is provided.}
\label{fig:arch_spatialet}
\end{figure}

%% ---------- Pretrained baseline ----------

%% file: supp_classes.tex
\section{Full 13-class frequency table}
\label{app:class_table}

\begin{table}[H]
\centering
\small
\begin{tabular}{@{}llr@{}}
\toprule
\textbf{Label} & \textbf{Meaning} & \textbf{Share (\%)} \\
\midrule
value          & Observed data cell (non-formula)       & 48.99 \\
aggregation    & Formula-derived (totals, subtotals)    &  0.91 \\
header         & Section/sheet titles and captions       &  0.07 \\
metadata       & Footnotes, sources, disclaimers         &  0.33 \\
comment        & Cell-stored note                        &  0.04 \\
empty          & Intentional blank for layout            & 45.44 \\
junk           & Placeholder text to exclude             &  0.23 \\
col\_header\_1 & Column header, innermost                &  0.70 \\
col\_header\_2 & Column header, mid level                &  0.07 \\
col\_header\_3 & Column header, outer level              &  0.01 \\
row\_header\_1 & Row header, innermost                   &  2.66 \\
row\_header\_2 & Row header, mid level                   &  0.33 \\
row\_header\_3 & Row header, outer level                 &  0.24 \\
\bottomrule
\end{tabular}
\caption{Empirical class distribution (${\sim}1$\,M labelled cells).}
\end{table}